\pdfoutput=1

\documentclass[11pt]{article}

\usepackage[preprint]{acl}

\usepackage{times}
\usepackage{latexsym}

\usepackage[T1]{fontenc}

\usepackage[utf8]{inputenc}

\usepackage{microtype}

\usepackage{inconsolata}
\usepackage{multirow}
\usepackage{pifont}%
\usepackage{makecell}
\usepackage[skins]{tcolorbox}
\tcbuselibrary{breakable}
\usepackage{algorithm}
\usepackage{algorithmic}
\usepackage[switch]{lineno}
\usepackage{booktabs}
\usepackage{graphicx}
\usepackage{amssymb}
\usepackage{enumitem}
\usepackage{amsmath}

\def\degree{${}^{\circ}$}
\newcommand{\cmark}{{\color{black}\ding{51}}}
\newcommand{\xmark}{{\color{gray}\ding{55}}}

\title{CorNav: Autonomous Agent with Self-Corrected Planning for Zero-Shot Vision-and-Language Navigation}

\author{%
Xiwen Liang$^1$$^*$, Liang Ma$^1$$^*$, Shanshan Guo$^2$, Jianhua Han$^3$, Hang Xu$^3$, \\
\textbf{Shikui Ma$^4$, Xiaodan Liang$^1$$^\dagger$} \\
$^1$Shenzhen Campus of Sun Yat-Sen University, $^2$Northeastern University, \\
$^3$Huawei Noah's Ark Lab, $^4$Dataa Robotics \\
\url{https://mligg23.github.io/MO-VLN-Site/}
}

\begin{document}

\maketitle

\let\thefootnote\relax\footnotetext{$^*$Equal contribution.}
\footnotetext{$^\dagger$Corresponding author.}

\begin{abstract}
Understanding and following natural language instructions while navigating through complex, real-world environments poses a significant challenge for general-purpose robots. These environments often include obstacles and pedestrians, making it essential for autonomous agents to possess the capability of self-corrected planning to adjust their actions based on feedback from the surroundings.
However, the majority of existing vision-and-language navigation (VLN) methods primarily operate in less realistic simulator settings and do not incorporate environmental feedback into their decision-making processes.
To address this gap, we introduce a novel zero-shot framework called CorNav, utilizing a large language model for decision-making and comprising two key components: 1) incorporating environmental feedback for refining future plans and adjusting its actions,
and 2) multiple domain experts for parsing instructions, scene understanding, and refining predicted actions.
In addition to the framework, we develop a 3D simulator that renders realistic scenarios using Unreal Engine 5. To evaluate the effectiveness and generalization of navigation agents in a zero-shot multi-task setting, we create a benchmark called NavBench.
Extensive experiments demonstrate that CorNav consistently outperforms all baselines by a significant margin across all tasks. On average, CorNav achieves a success rate of 28.1\%, surpassing the best baseline's performance of 20.5\%.

\end{abstract}    
\section{Introduction}
\label{sec:intro}

\begin{figure}
    \centering
    \includegraphics[width=\linewidth]{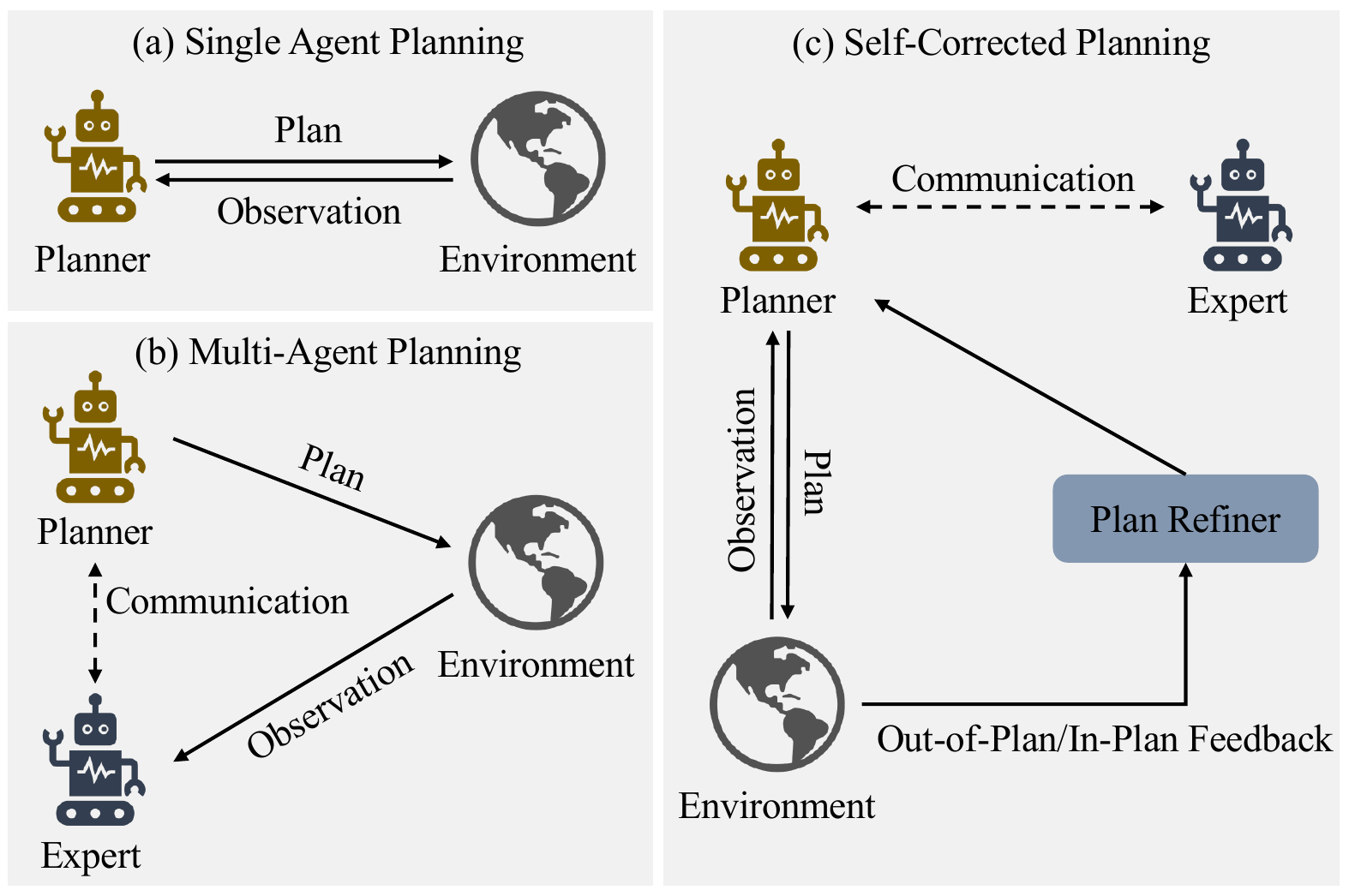}
    \vspace{-6mm}
    \caption{Comparison with existing VLN agents. (a) The single agent planning paradigm requires the agent to analyse and make decisions by itself. (b) Multi-agent planning paradigm enables the agent to communicate with multiple experts and perform complex reasoning. (c) Our self-corrected planning considers in-plan or out-of-plan feedback from a near-realistic environment.}
    \vspace{-8mm}
    \label{fig:compare_method}
\end{figure}

Language-driven navigation is a critical skill for robot assistants when it comes to performing a wide range of real-world tasks. Most autonomous agents are trained using predefined datasets and tasks, and perform well in familiar environments. However, the real world is filled with a multitude of objects and scenes, making it challenging to train an agent that can generalize effectively.
Recently, large language models (LLMs)~\cite{chowdhery2022palm,OpenAI2023GPT4TR,touvron2023llama,vicuna2023,koala_blogpost_2023} have demonstrated remarkable effectiveness across various tasks and have emerged as versatile autonomous agents capable of informed decision-making~\cite{sun2023adaplanner,wang2023survey,sumers2023cognitive}. These LLMs are pre-trained on massive textual data, endowing them with extensive commonsense knowledge that proves invaluable for navigation tasks. For instance, they can infer that a stove is likely to be found in the kitchen and that a bed is typically located in a bedroom.

The success of GPT has highlighted the efficacy of utilizing human instructions for zero-shot navigation tasks. Recently, zero-shot agents based on GPT~\cite{zhou2023navgpt,long2023discuss}, have harnessed the power of GPT-4~\cite{OpenAI2023GPT4TR} to make decisions in R2R dataset~\cite{anderson2018vision}. However, R2R dataset is based on a static, discrete, and unrealistic environment that lacks the complexity of real-world scenarios, including obstacle avoidance. These GPT-based methods may struggle when applied to real-world settings due to their limited consideration of environmental feedback.
Additionally, other agents~\cite{rajvanshi2023saynav,yu2023co} only focus on object navigation but are unable to comprehend complex instructions.

To address the aforementioned challenges, we present CorNav, an autonomous agent with self-corrected planning for zero-shot vision-and-language navigation in continuous environment. CorNav possesses several key capabilities, including the ability to understand complex instructions, engage in self-corrected planning based on both environmental and historical feedback, and consult domain experts for crucial information.
Here is a breakdown of CorNav's functionalities: 1) \textbf{Self-Corrected Planning:} During exploration, CorNav actively adapts its plan based on feedback. If the agent receives in-plan feedback, indicating that the environmental observation aligns with the plan, it adheres to the generated plan and proceeds with the next action. However, when faced with out-of-plan feedback, it modifies the plan accordingly. 2) \textbf{Domain Expert Consultation:} CorNav possesses complex reasoning and more accurate planning by seeking guidance from various domain experts. To manage computing resources and costs effectively, we have incorporated two key experts in addition to the vision perception expert: instruction parsing expert for understanding instructions and decision-making expert for evaluating and verifing the predicted actions.
The distinctive features and differences between CorNav and existing VLN agents are illustrated in Figure~\ref{fig:compare_method}.
Through a series of extensive experiments across multiple tasks, our agent has demonstrated outstanding performance, underscoring the effectiveness of its self-corrected planning mechanism and its ability to communicate and collaborate with multiple domain experts.

We also develop a near-realistic simulator using Unreal Engine 5\footnote{\url{https://www.unrealengine.com/}}, which offers enhanced lighting and intricate details compared to previous simulators~\cite{duan2022survey}. Our simulator encompasses four scenes carefully modeled from real-world scenarios.
Building upon this novel simulator, we have established a multi-task benchmark named NavBench for zero-shot vision-and-language navigation.
Unlike traditional data collection methods, we have harnessed the capabilities of powerful GPT-4 to generate high-quality instructions for various tasks.
NavBench has been designed to reflect realistic scenarios, covering four distinct tasks: 1) object navigation, namely, goal-conditioned navigation given a specific object category, which is a well-explored aspect of zero-shot navigation; 2) goal-conditioned navigation given a simple instruction, e.g., ``I want to go upstairs. Please help me find the elevator''; 3) completing abstract instruction, e.g., ``The floor is dirty and I want to sweep it'', implying that the robot should locate a broom; 4) step-by-step instruction following, which simulates common realistic navigation scenarios where the agent must follow a series of step-by-step instructions.
We have conducted an extensive study involving various large language models and open-vocabulary models, implementing 7 zero-shot baselines within the NavBench framework. Our experiments not only showcase the zero-shot capabilities of these foundational models but also highlight the challenging nature of NavBench as a benchmark.

In summary, our work presents three-fold contributions:
\begin{itemize}[topsep=-2pt,leftmargin=8pt]
\setlength\itemsep{-0.3em}
    \item \textbf{CorNav:} We introduce CorNav, a novel zero-shot VLN agent that stands out for its ability to adapt plans based on environmental feedback and its capacity to discuss with domain experts.
    \item \textbf{Realistic Simulator:} We have developed a realistic simulator using Unreal Engine 5, which provides a more immersive and challenging environment for our research.
    \item \textbf{NavBench:} We have established the NavBench benchmark, which leverages GPT-4 to generate and refine instructions in the dataset, eliminating the need for labor-intensive data collection.
\end{itemize}
\section{Related Work}
\label{sec:related_work}

\textbf{Vision-and-Language Navigation}
Language-guided visual navigation tasks have been a significant focus in recent research, and various models and tasks have contributed to advancing the field. The indoor navigation tasks such as R2R~\cite{anderson2018vision} and RxR~\cite{ku2020room} provide a foundation for language-guided navigation in simulated indoor environments. Many research efforts have been built upon these tasks, emphasizing cross-modal learning~\cite{ma2019self}, data augmentation~\cite{fu2020counterfactual,tan2019learning,liang2022contrastive}, waypoint tracking~\cite{deng2020evolving,ma2019regretful}, and pre-training using Transformer models to improve navigation performance~\cite{hao2020towards,hong2020language,liang2022visual}.
In addition, there are other tasks including outdoor navigation task Touch-Down \cite{chen2019touchdown}, dialogue-based navigation task CVDN \cite{thomason2020vision}, and remote object-grounded navigation, such as REVERIE \cite{qi2020reverie} and SOON \cite{zhu2021soon}, introducing the challenge of associating time-correlated visual observations with decision-making instructions. 

The performance of existing VLN methods often falls short when applied to the challenges of continuous 3D simulated environments, as exemplified by the more demanding task of VLN-CE \cite{krantz2020beyond}.
Recent advancements in the field, driven by the availability of large-scale datasets and the development of continuous environment simulators like Habitat \cite{savva2019habitat}, GibsonEnv \cite{eftekhar2021omnidata}, and AI2THOR \cite{kolve2017ai2}), have enabled a new set of tasks and benchmarks, which include PointGoal navigation \cite{wijmans2019dd,ye2021auxiliary}, ObjectGoal navigation \cite{chaplot2020learning,chaplot2020object,gervet2022navigating,ramakrishnan2022poni}, and instructions following navigation \cite{krantz2021waypoint,raychaudhuri2021language,hong2022bridging,an2022bevbert}.
Wang et al. \cite{wang2022towards} propose a large-scale indoor dataset designed for multimodal and multitask navigation in continuous and audiovisual complex environments. 

\textbf{Zero-shot Navigation}
The recent paradigm shifts in machine learning, driven by advancements in large-scale pre-training models, have indeed opened up exciting possibilities for zero-shot learning and have led to notable improvements in various downstream vision-language tasks, as demonstrated by Radford et al. \cite{radford2021learning}. In zero-shot navigation, CoW \cite{gadre2023cows} leverages CLIP for localization and frontier-based exploration (FBE) for exploration strategies. Dorbala et al. \cite{dorbala2022clip,dorbala2023can} subsequently used a Costmap to handle obstacle avoidance. ESC \cite{zhou2023esc} leverages a prompt-based language-image grounding model for open-world scene understanding and harnesses LLMs to acquire commonsense knowledge at object and room levels. Inspired by  recent advancements and the progress in open vocabulary \cite{radford2021learning,li2022grounded,kirillov2023segment,liu2023grounding,kamath2021mdetr} pre-training models, our work aims to empower embodied robots for improved navigation to uncommon objects.

\textbf{Large Language Model}
Large Language Models (LLMs) \cite{chowdhery2022palm,OpenAI2023GPT4TR,touvron2023llama,vicuna2023,koala_blogpost_2023,alpaca,chatGLM2022,bai2022constitutional,ouyang2022training,alayrac2022flamingo} have ushered in a significant transformation in the field of Artificial Intelligence (AI), particularly in language understanding, generation, and logical reasoning.
These models have evolved over the years, with recent breakthroughs primarily attributed to factors such as larger model sizes, enhanced pre-training data, instruction-following Tuning, and reinforcement learning from human feedback (RLHF) \cite{ouyang2022training,OpenAI2023GPT4TR,bai2022constitutional}.
Moreover, the development of hierarchical prompting systems for LLMs has gained prominence, aiming to enhance their logical reasoning abilities and response accuracy in specific domains \cite {wei2022chain,wang2022self,yao2022react,shinn2023reflexion,yao2023tree}.

\section{CorNav}

\begin{figure*}
    \centering
    \includegraphics[width=\linewidth]{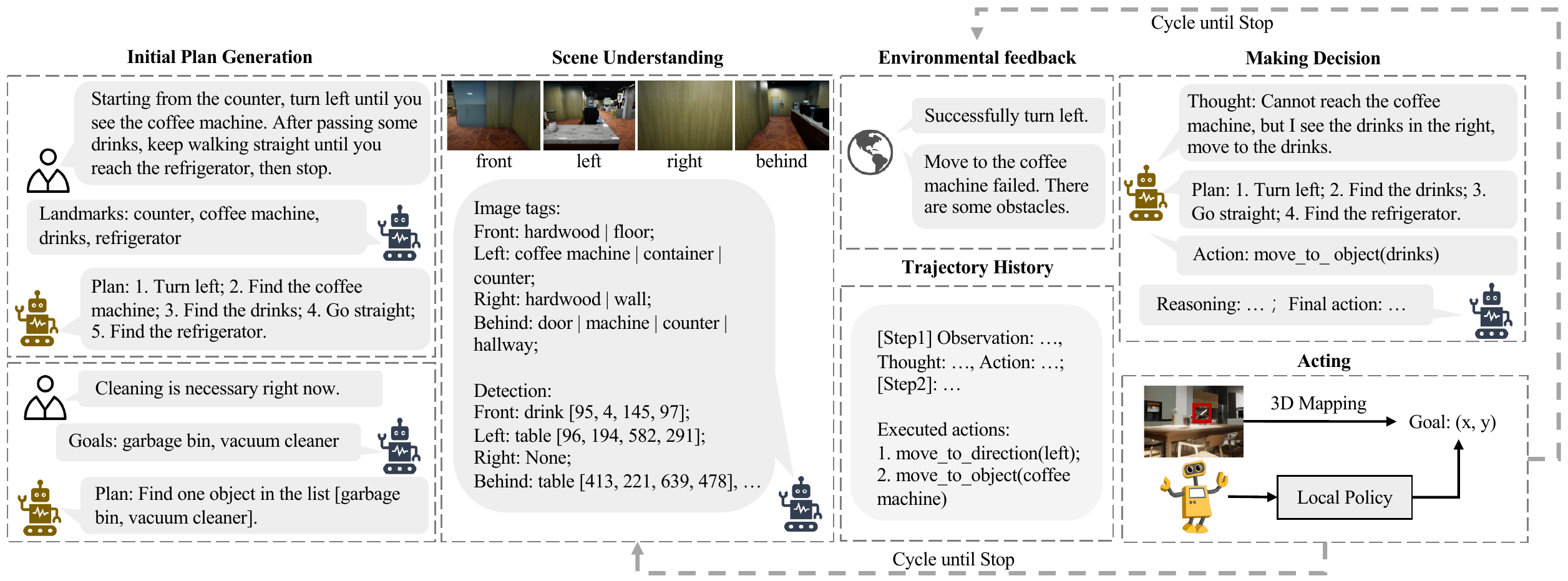}
    \vspace{-6mm}
    \caption{The overall architecture of our CorNav. After receiving the instruction, the instruction parsing expert extracts landmarks or figures out the needed objects. Then the agent generates the initial plan based on the instruction and information from the instruction parsing expert. The vision perception expert is driven by an image tagging model and an open-vocabulary grounding model, and performs scene understanding given four perspectives. Environmental feedback records both in-plan and out-of-plan feedback, while trajectory history maintains the reasoning process and executed actions. The decision-making expert assists the agent in deciding the final action. Finally, the local policy would plan a path for the robot.}
    \vspace{-6mm}
    \label{fig:architecture}
\end{figure*}

Our CorNav mainly comprises two pivotal components, i.e., multiple domain experts and plan refinement with environmental feedback. An overview of our architecture is shown in Figure~\ref{fig:architecture}. In this section, we will delve into the intricacies of these components, followed by a discussion of our navigation discussion mechanism and local policy.

\subsection{Domain experts}
Domain experts in our framework are powered by large models.  In this section, we introduce three core experts: the instruction parsing expert, the vision perception expert, and the decision-making expert. For the purpose of cost-efficiency, we have implemented the instruction parsing expert and decision-making expert using the open-source Large Language Model (LLM) Vicuna v1.5~\cite{vicuna2023}. Additionally, we have conducted comparisons with GPT-4 Turbo on a subset of our experiments to further evaluate performance.

\noindent
\textbf{Instruction Parsing Expert.}
Our benchmark encompasses a diverse range of navigation tasks, spanning from simple to complex instructions.
The instruction parsing expert excels at handling these instructions, particularly those of a more intricate nature.
For complex instructions, the role of the expert is to extract crucial information and deduce the intended goal objects.
In scenarios involving step-by-step instruction following, where the instruction may span several steps, the task of the expert is to extract relevant landmarks.
This process is depicted in the top left of Figure~\ref{fig:architecture}. Utilizing the original instruction and the extracted landmarks, the planner agent generates an initial navigation plan, setting the stage for successful execution.
In tasks such as completing abstract instructions, where specific object names may not be explicitly mentioned, the instruction parsing expert steps in to decipher the intent of the user. For instance, when presented with the instruction ``I am thirsty'', the expert deduces that the user likely requires water or a beverage and produces a list of potential options, such as ``[water, drink]''. This capability is showcased in the bottom left of Figure~\ref{fig:architecture}.

\noindent
\textbf{Vision Perception Expert.}
Vision perception is a fundamental module in VLN, tasked with providing comprehensive and accurate environmental information to aid navigation. To achieve this, we integrate two key components into this expert: an image tagging model and an open-vocabulary grounding model. For image tagging, we employ the robust and state-of-the-art model RAM++~\cite{huang2023inject}. The vision perception expert processes input images captured from four different perspectives of the robot: front, left, right, and rear. RAM++ then predicts object tags for each of these perspectives. It is important to note that occasional noise in the predicted results can disrupt the planner's decision-making process. To mitigate the potential impact of noise in the predicted results, we introduce an open-vocabulary grounding model Grounding DINO~\cite{liu2023grounding} for object detection. This model prompts the planner to pay more attention on the objects that have been reliably detected.
By combining image tagging and object detection capabilities, the expert enhances th overall environmental perception for the agent.

\noindent
\textbf{Decision-making Expert.}
The decision-making expert plays a pivotal role in overseeing and refining the predicted actions.
The system employs four primary types of global actions, each serving a distinct purpose: move\_to\_object(), move\_to\_room(), move\_to\_direction(), and stop(). move\_to\_object($o$) signifies the intention to approach and navigate near object $o_i$, while move\_to\_room($r$) indicates the desire to enter and navigate into room $r$. If the agent faces uncertainty regarding object or room selection, then it would move in the most probable direction (i.e., front, left, right, rear) considering the available environmental information.
The primary responsibility of the expert is to review and validate the decisions generated by the planner. Vicuna v1.5 is not as intelligent as GPT-4, and might present a sub-optimal decision. For example, the planner thinks that ``Since the kettle is not detected in front of me, I will move to the left to check the shelf'', and gives the action ``move\_to\_direction(left)''. Given the thought from the planner, the decision-making expert should recognize that the kettle may be on the shelf, and modify the action to a more suitable one ``move\_to\_object(shelf)''.

\subsection{Environmental feedback}
\label{sec:feedback}
After generating the initial plan, the planner proceeds to predict the next global action based on this plan. However, real-world environments can introduce unexpected challenges, occasionally necessitating adjustments to the original plan to accommodate these changes.

\noindent
\textbf{In-Plan Feedback.}
In scenarios where the predicted action is move\_to\_object($o$) or move\_to\_room($r$), we leverage the detection model Grounding DINO to localize object $o$ or room $r$. Then we can obtain the location of $o$ or $r$ in the simulator by transforming pixels in 2D images into 3D voxels, taking into account the agent's location, camera perspective, and depth information. To obtain more precise object locations, we employ SAM~\cite{kirillov2023segment} for semantic segmentation. Then we can calculate the distance between the  the agent's stopping location and the detected location of $o$ or $r$.
If this distance falls below a predefined threshold, we consider the action successful, indicating that the agent has effectively reached the vicinity of the object or room. Conversely, if the distance exceeds the threshold, the action is considered a failure.
When the agent successfully moves near object $o$ or room $r$, it signifies that the environment aligns with the anticipated plan. In such cases, the in-plan feedback is documented as ``successfully move to $o$ or $r$''.
For predicted actions of move\_to\_direction($d$), the agent is expected to traverse a specific distance in the specified direction. Analogous to the previous cases, if the agent accomplishes this task without issues, the in-plan feedback is registered as ``successfully turn to direction $d$''.
When the feedback indicates success, the planner retains the previous plan, as it accurately corresponds to the environment.

\noindent
\textbf{Out-of-Plan Feedback.}
As outlined earlier, our system provides a means to determine whether the agent successfully executes a planned action.
When an action is executed unsuccessfully, the system registers out-of-plan feedback, indicating that obstacles or challenges were encountered during the execution. 
If the agent's attempt to move to object $o$ or room $r$ results in failure, it is indicated as ``move to $o$ or $r$ failed''. This feedback suggests that obstacles or impediments prevented the successful execution of the action, requiring further adaptation. Similarly, for actions involving move\_to\_direction($d$), a failure is noted as ``turn to direction $d$ failed''.
Let $\{ a_1, a_2, ..., a_t \}$ be the original plan. Upon receiving out-of-plan feedback at a particular step $i$, the planner takes corrective action. Specifically, the planner discards the subsequent actions in the original plan, resulting in the formulation of a new plan $\{ a_1, a_2, ..., a_{i-1}, a'_i, ..., a'_t \}$ that considers the environmental feedback and observations.
This revised plan is designed to guide the planner in making informed and adaptive decisions.

\subsection{Navigation Discussion Mechanism}
In this part, we delve into the comprehensive navigation discussion process, depicted in Figure~\ref{fig:architecture}.
Initially, the planner calls upon the instruction parsing expert $\mathcal{I}$ to extract landmarks or infer goals from the instruction $I$. Subsequently, the planner generates an initial plan $p$ based on the instruction and insights provided by the instruction parsing expert:
\begin{equation}
    p = \mathcal{P}(I, \mathcal{I}(I)).
\end{equation}
Simultaneously, the vision perception expert $\mathcal{V}$ processes observations $\mathcal{O}$ received from the environment. This expert summarizes the outcomes of image tagging and object detection. The trajectory history buffer $\mathcal{H}$ serves as a repository of historical information, encompassing observations, thoughts, and executed actions (${a}$).
At each time step $t$, if the planner receives out-of-plan feedback $f$, it triggers a series of actions to generate a new plan and decide as follows:
\begin{equation}
\begin{aligned}
    p' = \mathcal{P}(\mathcal{H}, \mathcal{V}(\mathcal{O}), I, \mathcal{I}(I), p, \{a\}, f), \\
    T_{t+1}, a_{t+1} = \mathcal{P}(\mathcal{H}, \mathcal{V}(\mathcal{O}), I, \mathcal{I}(I), p', \{a\}),
\end{aligned}
\end{equation}
where $T_{t+1}$ indicates thought for taking action $a_{t+1}$.
In the absence of out-of-plan feedback, the planner relies on the original plan $p$ for decision-making. Subsequently, the planner engages the decision-making expert $\mathcal{D}$ to arrive at a final decision:
\begin{equation}
    T'_{t+1}, a'_{t+1} = \mathcal{D}(T_{t+1}, a_{t+1}).
\end{equation}

\subsection{Local Policy}
\label{sec:local_policy}
Once the agent has ascertained the goal location, as outlined in Section~\ref{sec:feedback}, the local policy engages the Fast Marching Method \cite{sethian1996fast} to formulate a path from the current location to the designated goal. This path planning process ensures efficient and effective navigation to reach the specified objective.
\section{NavBench}

\begin{table*}[t]
    \centering
    \caption{Comparison with embodied AI simulators. Physics simulation: basic physics features (B) and advanced physics features (A). Model library support: built-in (L) and user-extensible (E). Action interactivity: navigation (N), object manipulation (M), and human-computer interaction using virtual reality (VR) devices (H). Pedestrian: adding walking and stationary pedestrians. Visual quality: 5 indicates most realistic, while 1 represents least.}
    \vspace{-2mm}
    \resizebox{\linewidth}{!}{
    \begin{tabular}{c|c|c|c|c|c|c|c|c}
        \toprule
        Simulator & Simulation Engine & Physics & Models & Action & Pedestrian & \makecell{Object \\ number} & \makecell{Object \\ category} & \makecell{Visual \\ quality} \\
        \midrule
        DeepMind Lab \cite{beattie2016deepmind} & Quake II Arena Engine & - & - & N & \xmark & - & - & 1.6 \\
        CHALET \cite{yan2018chalet} & Unity 3D Engine & B & - & N, M & \xmark & 1740 & 150 & 2.7 \\
        VirtualHome \cite{puig2018virtualhome} & Unity 3D Engine & - & - & N, M, H & \xmark & 2142 & 308 & 2.5 \\
        VRKitchen \cite{gao2019vrkitchen}] & Unreal Engine 4 & B & - & N, M & \xmark & 880 & - & 2.4 \\
        Habitat-Sim \cite{savva2019habitat} & Bullet & - & - & N & \xmark  & 92 & - & 2.2 \\
        AI2-THOR \cite{kolve2017ai2} & Unity 3D Engine & B & L & N, M & \xmark & 609 & - & 3.2 \\
        iGibson \cite{xia2020interactive} & PyBullet & B & L & N, M & \xmark & 570 & - & 2.8 \\
        SAPIEN \cite{xiang2020sapien} & PhysX Physical engine and ROS & B & L & N, M & \xmark & 2346 & 46 & 1.8 \\
        ThreeDWorld \cite{gan2020threedworld} & Unity 3D Engine &  B, A & L, E & N, M, H & \xmark & 2500 & 200 & 3.2 \\
        BEHAVIOR-1K \cite{li2023behavior} & Nvidia’s Omniverse & B, A & L, E & N, M & \xmark & 5215 & 1265 & 3.5 \\
        \midrule
        NavBench (Ours) & Unreal Engine 5 & B, A & L, E & N, M, H & \cmark & 4758 & 2165 & 4.3 \\
        \bottomrule
    \end{tabular}}
    \label{tab:compare_sim}
    \vspace{-2mm}
\end{table*}

Our simulator is featured for indoor navigation. It stands out by seamlessly integrating a state-of-the-art real-time physics engine, which significantly elevates the quality of visual rendering. Notably, our simulator boasts dynamic global illumination and diffuse global illumination, allowing for more highly detailed geometry rendering than ever before.
Table~\ref{tab:compare_sim} offers a comprehensive comparison between our simulator and existing counterparts, highlighting the distinctive features and capabilities that set ours apart.
We conduct a visual quality assessment involving 60 human participants on previous and our proposed simulators in Table \ref{tab:compare_sim}.

Our benchmark, NavBench, is purposefully designed to address the challenges of zero-shot multi-task vision-and-language navigation. It encompasses a wide range of tasks, including navigation to specific goal objects, abstract objects, and specific locations, all guided by natural language instructions. Unlike previous benchmarks relying on manual data labeling, we collect data using large language models. Table~\ref{tab:compare_benchmark} succinctly outlines the key distinctions between NavBench and previous benchmarks.

\begin{table*}[]
    \centering
    \caption{Comparison with existing vision-and-language benchmarks.}
    \vspace{-2mm}
    \resizebox{\linewidth}{!}{
    \begin{tabular}{c|cccc}
        \toprule
        Benchmark & Simulator & Continuous & Tasks & Instruction Type \\
        \midrule
        R2R \cite{anderson2018vision} & Matterport3D \cite{anderson2018vision} & \xmark & 1 & Route-oriented \\
        RoomNav \cite{wu2018building} & House3D \cite{wu2018building} & \cmark & 1 & Goal-oriented \\
        LANI \cite{misra2018mapping} & AI2-THOR \cite{kolve2017ai2} & \cmark & 1 & Goal-oriented \\
        3D Doom \cite{chaplot2018gated} & VizDoom \cite{kempka2016vizdoom} & \cmark & 1 & Goal-oriented \\
        VNLA \cite{nguyen2019vision} & Matterport3D & \xmark & 1 & Oracle guidance \\
        HANNA \cite{nguyen2019help} & Matterport3D & \xmark & 1 & Oracle guidance \\
        R4R \cite{jain2019stay} & Matterport3D & \xmark & 1 & Route-oriented \\
        CVDN \cite{thomason2020vision} & Matterport3D & \xmark & 1 & Dialogue \\
        R6R, R8R \cite{zhu2020babywalk} & Matterport3D & \xmark & 1 & Route-oriented \\
        RxR \cite{ku2020room} & Matterport3D & \xmark & 1 & Route-oriented \\
        VLNCE \cite{krantz2020beyond} & Matterport3D & \xmark & 1 & Route-oriented \\
        REVERIE \cite{qi2020reverie} & Matterport3D & \xmark & 1 & Goal-oriented \\
        SOON \cite{zhu2021soon} & Matterport3D & \xmark & 1 & Goal-oriented \\
        BnB\cite{guhur2021airbert} & - & \xmark & 1 & Route-oriented \\
        ROBUSTNAV \cite{chattopadhyay2021robustnav} & ROBOTHOR \cite{deitke2020robothor} & \cmark & 2 & Goal-oriented \\
        PASTURE \cite{gadre2023cows} & ROBOTHOR & \cmark & 3 & Goal-oriented \\
        \midrule
        NavBench (Ours) & Ours & \cmark & 4 & Goal-oriented, Route-oriented \\
        \bottomrule
    \end{tabular}}
    \label{tab:compare_benchmark}
    \vspace{-5mm}
\end{table*}

\subsection{Task Definition}
To construct our benchmark, we have categorized it into four distinct tasks, including object navigation given a category, goal-conditioned navigation given a simple instruction, completing abstract instruction, and step-by-step instruction following.

\noindent
\textbf{Object navigation given a category (ObjectNav).}
This task revolves around the navigation to specific objects based on their predefined categories within our simulated environments.

\noindent
\textbf{Goal-conditioned navigation given a simple instruction (Simple).}
In this task, the language instructions provided to the agent contain references to object categories or closely related terms. For example, instructions generated by GPT-4 may resemble, ``Proceed toward the nearest mug that is detectable'', where the term ``mug'' is included in the instruction.

\noindent
\textbf{Completing abstract instruction (Abstract).}
Here, the instructions issued to the agent are intentionally abstract and do not explicitly mention object names. Instead, the agent must infer the intended goal based on the user's abstract intent. For instance, when presented with the instruction ``I am thirsty'', the agent should deduce that the user requires water or a beverage and output a list such as ``[water, drink]''.

\noindent
\textbf{Step-by-step instruction following (Step-by-step).}
In this task, the agent is required to follow detailed step-by-step instructions provided in the language. This mirrors real-world navigation scenarios where complex instructions guide the agent's actions.

It is noteworthy that for the last three tasks, we employ the cutting-edge GPT-4 \cite{OpenAI2023GPT4TR} to generate the dataset. GPT-4's remarkable language generation capabilities are instrumental in crafting realistic and diverse instructions for these tasks, enabling a comprehensive evaluation of the agent's performance.

\subsection{Dataset Statistics}
Our benchmark encompasses four distinct scenes, including a restaurant, cafe, nursing room, and home settings.
In total, our dataset comprises a substantial corpus of 1,615 instructions. Specifically, the distribution of instructions across tasks is as follows:
81 instructions for ObjectNav, 494 instructions for the simple tasks, 278 for the abstract task, and 762 for the step-by-step task.
More details are shown in Appendix.
\section{Experiment}

\subsection{Experimental Setup}
\textbf{Navigation metrics.}
We use standard navigation metrics to measure performance: Success Rate (SR), the fraction of episodes where the agent successfully reaches within 1.5m of the target object or location; Success Rate weighted by inverse path Length (SPL), success weighted by the oracle shortest path length and normalized by the actual path length \cite{batra2020objectnav}; and Distance to Success (DTS), the distance of the agent from the success threshold boundary when the episode ends~\cite{chaplot2020object}.

\noindent
\textbf{Embodiment.}
We define four actions: \emph{Move Forward}, \emph{Turn Left}, \emph{Turn Right}, and \emph{Stop}. The \emph{Move Forward} action advances the agent by 20cm, while \emph{Turn Left} and \emph{Turn Right} actions turn 15\degree 
horizontally.

\subsection{Comparison of Zero-Shot Methods}
We have implemented a total of seven baseline models, leveraging three distinct methods for zero-shot object navigation in a continuous environment, as detailed in Table \ref{tab:compare_baselines}. Recognizing that these methods are primarily designed for object navigation and may struggle with longer instructions, we employ LLMs to parse instructions, which is the same as our instruction parsing expert.
The results of these baselines are summarized in Table \ref{tab:compare_baselines}. Notably, the models incorporating GLIP or Grounding DINO tend to outperform the CoW baseline. Interestingly, the co-occurrence knowledge from LLMs in ESC appears to have a lesser impact on the results. It is worth noting that selecting boundaries based on common sense may not be ideal in our specific scenarios.

\subsection{Compare CorNav with Previous Methods}
In our evaluation, we compare CorNav with previous methods, and the results are summarized in Table \ref{tab:compare_baselines}. Notably, our method outperforms all baselines across all four tasks, achieving an average Success Rate (SR) of 28.1\%. This represents a significant improvement, with a 7.6\% increase compared to the best-performing baseline.
Particularly noteworthy is CorNav's remarkable performance in the step-by-step task, where it achieves an 8.6\% increase in SR. This outcome underscores the effectiveness of our approach, which incorporates environmental feedback and leverages trajectory history to enhance navigation capabilities.

\begin{table*}[t]
    \centering
    \caption{Multi-task navigation results on NavBench.}
    \vspace{-2mm}
    \resizebox{1.0\linewidth}{!}{
    \begin{tabular}{l|c|cc|cc|cc|cc|c}
        \toprule
        \multicolumn{2}{c|}{} & \multicolumn{2}{c|}{ObjectNav} & \multicolumn{2}{c|}{Simple} & \multicolumn{2}{c|}{Abstract} & \multicolumn{2}{c|}{Step-by-step} & Avg. \\
        \midrule
        Model & Detector & SR & SPL & SR & SPL & SR & SPL & SR & SPL & SR \\
        \midrule
        CoW~\cite{gadre2023cows} & CLIP-Grad. & 12.2 & 10.2 & 15.0 & 11.5 & 28.4 & 25.8 & 18.5 & 6.7 & 18.5 \\
        FBE~\cite{yamauchi1997frontier} & GLIP~\cite{li2022grounded} & 18.3 & 14.3 & 17.4 & 13.7 & 28.1 & 24.0 & 17.7 & 13.8 & 20.4 \\
        FBE~\cite{yamauchi1997frontier} & Grounding DINO~\cite{liu2023grounding} & 17.1 & 11.4 & 15.6 & 10.8 & 27.1 & 21.2 & 19.5 & 12.6 & 19.8 \\
        ESC (Vicuna v1.5-13B)~\cite{zhou2023esc} & GLIP & 15.8 & 12.9 & 14.8 & 11.6 & 25.6 & 22.5 & 17.7 & 13.5 & 18.5 \\
        ESC (Vicuna v1.5-13B)~\cite{zhou2023esc} & Grounding DINO & 18.3 & 11.3 & 15.2 & 9.2 & 25.9 & 20.1 & 20.8 & 11.2 & 20.1 \\
        ESC (GPT-4)~\cite{zhou2023esc} & GLIP & 15.9 & 13.3 & 16.8 & 13.2 & 26.0 & 23.6 & 20.5 & 16.2 & 19.8 \\
        ESC (GPT-4)~\cite{zhou2023esc} & Grounding DINO & 18.3 & 11.7 & 15.6 & 10.0 & 30.2 & 22.9 & 18.0 & 9.9 & 20.5 \\
        \midrule
        CorNav (Vicuna v1.5-13B) & Grounding DINO & \textbf{23.4} & \textbf{16.1} & \textbf{23.7} & \textbf{19.8} & \textbf{36.0} & \textbf{29.0} & \textbf{29.4} & \textbf{23.1} & \textbf{28.1} \\
        \bottomrule
    \end{tabular}}
    \label{tab:compare_baselines}
    \vspace{-4mm}
\end{table*}

\subsection{Ablation Study}

\textbf{The effect of environmental feedback.}
To assess the significance of environmental feedback, we conducted an ablation study, the results of which are presented in Table \ref{tab:objectnav}. Notably, the inclusion of the plan refiner with environmental feedback (row 2) yields a remarkable improvement over the baseline (row 1). This highlights the crucial role that environmental feedback plays in enhancing the realism of navigation.

\noindent
\textbf{The effect of trajectory history.}
Examining the impact of trajectory history, as demonstrated in Table \ref{tab:objectnav} reveals that planning with trajectory history (row 3) further improves the results compared to the plan refiner alone (row 2). This observation aligns with the inherent logic of navigation, where past actions and experiences inform future decisions.

\noindent
\textbf{The effect of multiple experts.}
Our study also delves into the effects of consulting with domain experts, specifically, the decision-making expert and the instruction parsing expert. As depicted in Table \ref{tab:objectnav}, involving the decision-making expert contributes to improved navigation outcomes, suggesting instances where the agent's decisions might have been sub-optimal. Further insights emerge from Table \ref{tab:instruction_parising}, where the instruction parsing expert exhibits significant enhancements in the abstract task (SR +8.3\%). Parsing instructions becomes particularly important in scenarios where object names are absent, emphasizing its importance.

\noindent
\textbf{The effect of environmental description.}
The vision perception expert incorporates both an image tagging model and an object detection model. An ablation study, detailed in Table \ref{tab:description}, reveals that while utilizing either image tags or detection results alone yields similar performance, combining both aspects results in significantly improved performance.

\begin{table}[t]
    \centering
    \caption{Ablation study about different components of CorNav on ObjectNav.}
    \vspace{-2mm}
    \resizebox{1.0\linewidth}{!}{
    \begin{tabular}{l|ccc}
        \toprule
        Method & SR & SPL & DTS (m) \\
        \midrule
        Baseline & 18.5 & 13.8 & 7.85 \\
        $+$ Environmental Feedback & 21.0 & \textbf{16.1} & 7.74 \\
        $+$ Trajectory History & 22.2 & 16.0 & 7.80 \\
        $+$ Decision-making Expert & \textbf{23.4} & \textbf{16.1} & \textbf{7.52} \\
        \bottomrule
    \end{tabular}}
    \label{tab:objectnav}
    \vspace{-3mm}
\end{table}

\begin{table}[t]
    \centering
    \caption{Ablation study about instruction parsing expert on complex tasks.}
    \vspace{-3mm}
    \resizebox{1.0\linewidth}{!}{
    \begin{tabular}{l|cc|cc}
        \toprule
         & \multicolumn{2}{c|}{Abstract} & \multicolumn{2}{c}{Step-by-step} \\
        \midrule
        Method & SR & SPL & SR & SPL \\
        \midrule
        w/o instruction parsing expert & 27.7 & 21.7 & 28.1 & 21.7 \\
        CorNav (Vicuna v1.5-13B) & \textbf{36.0} & \textbf{29.0} & \textbf{29.4} & \textbf{23.1} \\
        \bottomrule
    \end{tabular}}
    \label{tab:instruction_parising}
    \vspace{-5mm}
\end{table}

\begin{table}[t]
    \centering
    \caption{Comparison of results from different environmental description on ObjectNav.}
    \vspace{-2mm}
    \resizebox{1.0\linewidth}{!}{
    \begin{tabular}{l|ccc}
        \toprule
        Description & SR & SPL & DTS (m) \\
        \midrule
        Image Tags & 21.0 & 15.2 & 7.95 \\
        Detection & 21.0 & 14.6 & 8.29 \\
        Image Tags $+$ Detection & \textbf{23.4} & \textbf{16.1} & \textbf{7.52} \\
        \bottomrule
    \end{tabular}}
    \label{tab:description}
    \vspace{-3mm}
\end{table}

\begin{table}[t]
    \centering
    \caption{Comparison between Vicuna v1.5-13B and GPT-4 Turbo on a subset containing four tasks.}
    \vspace{-2mm}
    \resizebox{1.0\linewidth}{!}{
    \begin{tabular}{l|ccc}
        \toprule
        Method & SR & SPL & DTS (m) \\
        \midrule
        CorNav (Vicuna v1.5-13B) & 20.8 & 13.0 & 7.01 \\
        CorNav (GPT-4 Turbo) & \textbf{27.1} & \textbf{18.5} & \textbf{5.87} \\
        \bottomrule
    \end{tabular}}
    \label{tab:gpt4}
    \vspace{-6mm}
\end{table}

\noindent
\textbf{Compare between Vicuna v1.5 and GPT-4.}
We conducted a comparative analysis between Vicuna v1.5-13B and GPT-4 Turbo, focusing on a subset of our dataset. For this subset, we randomly selected three instructions from each scene for each task, resulting in a total of 48 instructions. The results are presented in Table \ref{tab:gpt4}.
Remarkably, GPT-4 Turbo exhibited a substantial improvement in performance (+6.3\% on SR) compared to Vicuna v1.5. This suggests that GPT-4 Turbo operates as a more intelligent agent.
\section{Conclusion}
In this paper, we introduce CorNav, an innovative autonomous agent designed for zero-shot VLN. CorNav excels in leveraging environmental feedback to refine its plans in realistic scenarios, ensuring adaptability to dynamic surroundings. It also incorporates multiple domain experts for instruction parsing, scene comprehension, and action refinement. Our experimental results demonstrate CorNav's significant performance advantages over baseline methods across various navigation tasks.
Furthermore, we contribute to the field by developing a more realistic simulator powered by Unreal Engine 5. To evaluate our agent's capabilities, we create NavBench, a comprehensive multi-task benchmark for open-set zero-shot VLN. Leveraging the powerful GPT-4, we generate and self-refine a range of free-form instructions for different tasks within NavBench, including goal-conditioned navigation, abstract object retrieval, and step-by-step instruction following. Our benchmark offers a challenging platform for assessing navigation methods.

\section*{Acknowledgements}
This work was supported in part by CAAI-Huawei MindSpore Open Fund. We also thank MindSpore\footnote{\url{https://www.mindspore.cn/}} for the partial support of this work, which is a new deep learning computing framework.

\bibliography{main}

\appendix

\section{More Method Details}

\begin{figure}[h]
    \centering
    \includegraphics[width=1.0\linewidth]{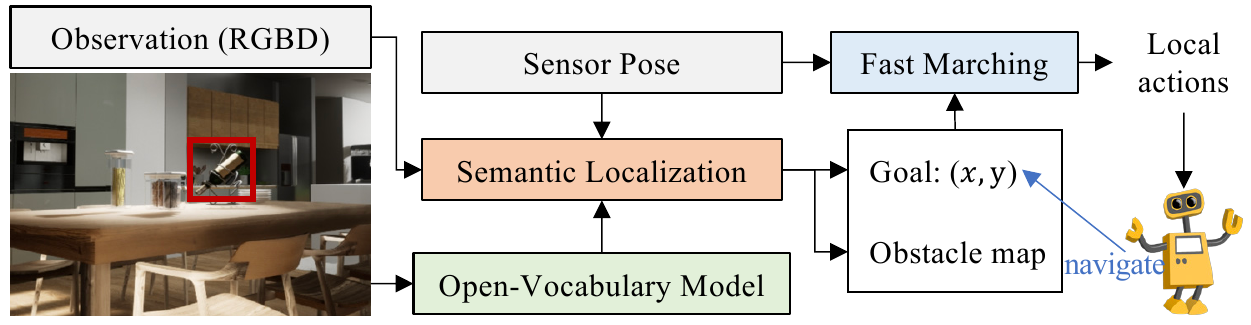}
    \vspace{-4mm}
    \caption{The illustration of the acting module in CorNav.}
    \vspace{-2mm}
    \label{fig:acting}
\end{figure}

In this section, we elaborate on the functionalities of the acting module, as depicted in Figure \ref{fig:acting}. In scenarios where the predicted action is move\_to\_object($o$) or move\_to\_room($r$), our approach incorporates the open-vocabulary model such as Grounding DINO for localizing the specified object $o$ or room $r$. Subsequently, the localization of $o$ or $r$ in the simulation environment is achieved by converting pixel representations in 2D image into 3D voxel. This conversion process takes into account the agent's location, the camera's perspective, and depth information. In order to enhance the precision of object localization, we engage the SAM~\cite{kirillov2023segment} for semantic segmentation in semantic localization. Following the acquisition of both the obstacle map and the designated goal location, the agent navigates towards the goal utilizing the fast marching technique as in Section \ref{sec:local_policy}.

\section{Simulator}

\begin{figure*}[t]
    \centering
    \includegraphics[width=1.0\linewidth]{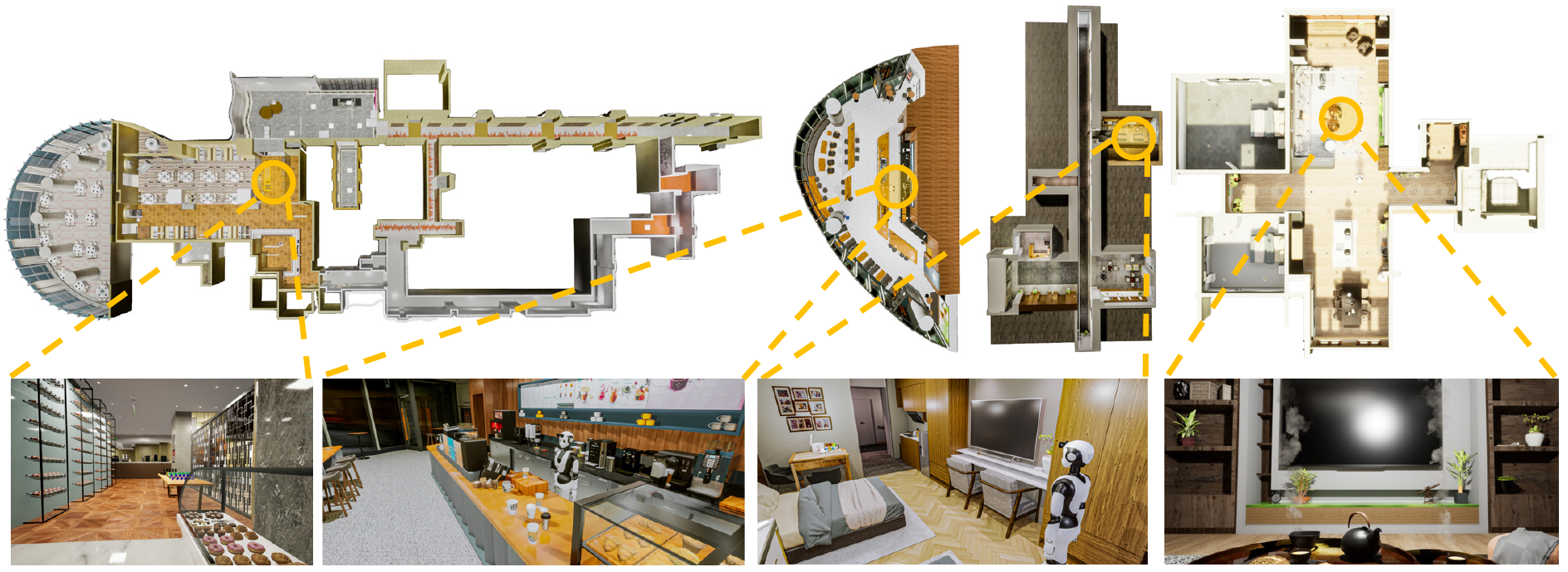}
    \vspace{-4mm}
    \caption{Our simulator includes scenes of different difficulty, i.e., restaurant, cafe, nursing room, and home.}
    \vspace{-2mm}
    \label{fig:scene}
\end{figure*}

\textbf{Scenes}
Our simulator provides four scenes, i.e., cafe, restaurant, nursing room, and home, which have a relatively large demand for robots. The overview of scenes is shown in Figure \ref{fig:scene}. These scenes are rendered based on real-world scenarios and contain more realistic details. The simulator is flexible and users can change the lighting and place more additional objects in the scene.

\begin{figure*}[t]
    \centering
    \includegraphics[width=1.0\linewidth]{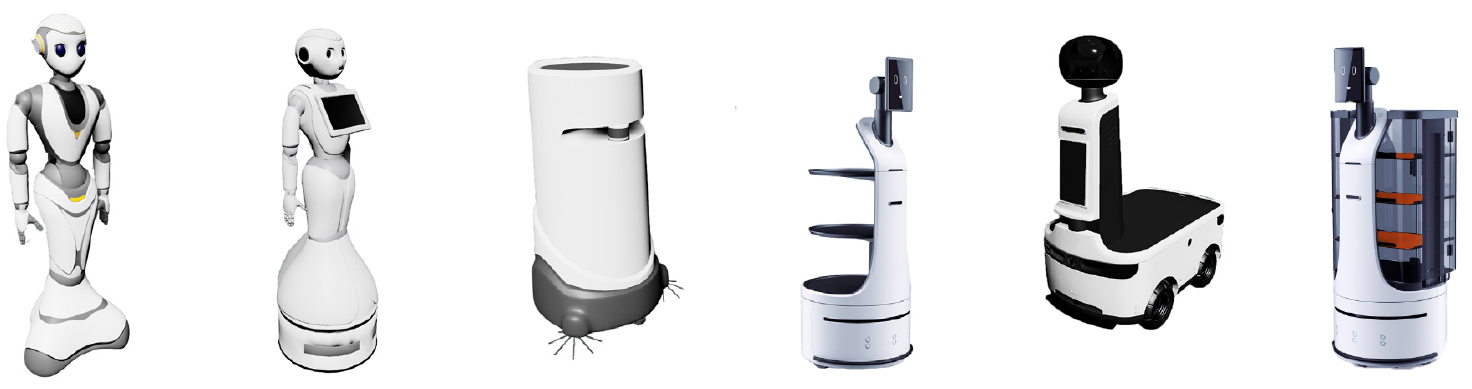}
    \vspace{-4mm}
    \caption{Supported agents in our simulator. We include agents in a variety of application scenarios, such as humanoid agents, sweeping agents, and delivery agents.}
    \vspace{-2mm}
    \label{fig:robot}
\end{figure*}

\textbf{Agents}
Our simulator supports multiple agents with different practical uses. For example, the humanoid robot can navigate like a human and perform more actions such as turning the head or nodding to have a wider view, while the sweeping robot aims at cleaning the floor. We list all supported agents in Figure \ref{fig:robot}. Different agents have different movable joints. We mainly use the humanoid agent in our experiments.

\textbf{Actions}
Our simulator supports continuous move or teleport actions. Users can define discrete actions such as rotating right by 30\degree. The humanoid agent has 21 movable joints that can make all human movements, including rotation of the head, neck, and waist.

\begin{figure*}[t]
    \centering
    \includegraphics[width=1.0\linewidth]{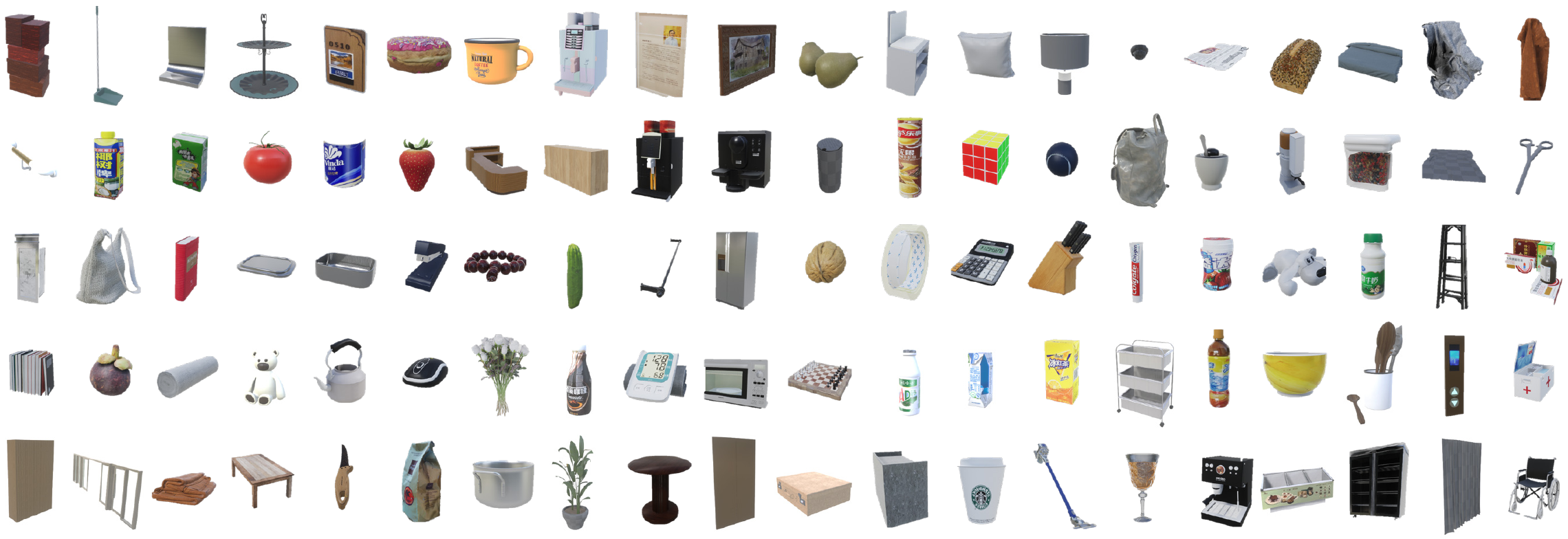}
    \vspace{-2mm}
    \caption{Examples of objects in our simulator.}
    \vspace{-2mm}
    \label{fig:objects}
\end{figure*}

\textbf{Objects}
Our simulator was built with 2,165 categories in total. We choose 129 categories among them for interaction.
Except for common objects in indoor environments, our simulator also includes some uncommon objects and more fine-grained categories, such as ``soft drink'' and ``juice''. Figure \ref{fig:objects} shows some samples of objects in our simulator. Users can generate more objects in the scene by Python API.

\subsection{Image Modalities}

\begin{figure*}[t]
    \centering
    \includegraphics[width=0.9\linewidth]{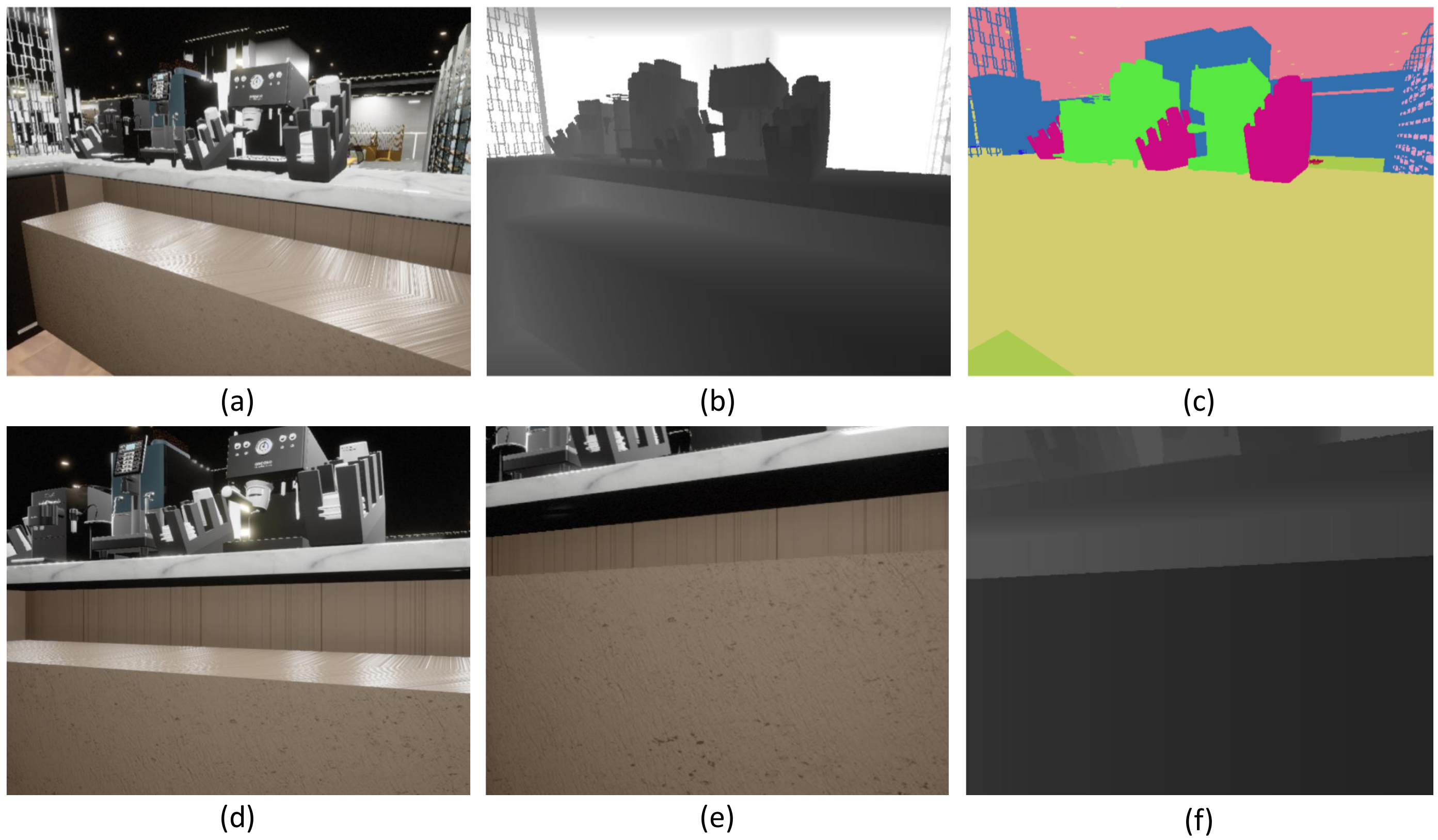}
    \vspace{-2mm}
    \caption{Examples of image modalities in our simulator: (a) RGB image from head camera; (b) depth image from head camera; (c) semantic segmentation mask from head camera; (d) RGB image from chest camera; (e) RGB image from waist camera; (f) depth image from waist camera.}
    \label{fig:image_modalities}
\end{figure*}

Different image modalities from different cameras are shown in Figure \ref{fig:image_modalities}. There are three image modalities in the scene, including RGB, depth, and semantic segmentation. Some humanoid agents have three cameras on the head, chest, and waist, respectively.

\subsection{3D Model}

Our simulator is built based on real scenarios at a 1:1 ratio. We use 689 object models, classified into 534 categories, to build the cafe scene. For building the restaurant scene, we employ 2,023 object models classified into 782 categories. For the nursing house scene, we adopt 1,518 object models, which can be classified into 849 categories. For the home scene, we utilize 528 models.

Since our ultimate goal is to develop intelligent robots that can perform multiple tasks such as navigation and grasping in the future, we select some of these categories to build the benchmark. We have chosen 65 categories in the four scenes to evaluate navigation approaches. Among these scenes, the cafe contains 17 categories, the restaurant includes 14 categories, the nursing room has 32 categories, and the home scene contains 18 categories.

We also provide Python API for users to generate more objects in the scene. For example, put more stuff on the table or on the floor. In this way, they can change the layout of objects in the scene by themselves. There are a total of 129 categories for interaction.

\subsection{Pedestrian}
\begin{figure*}[t]
    \centering
    \includegraphics[width=\linewidth]{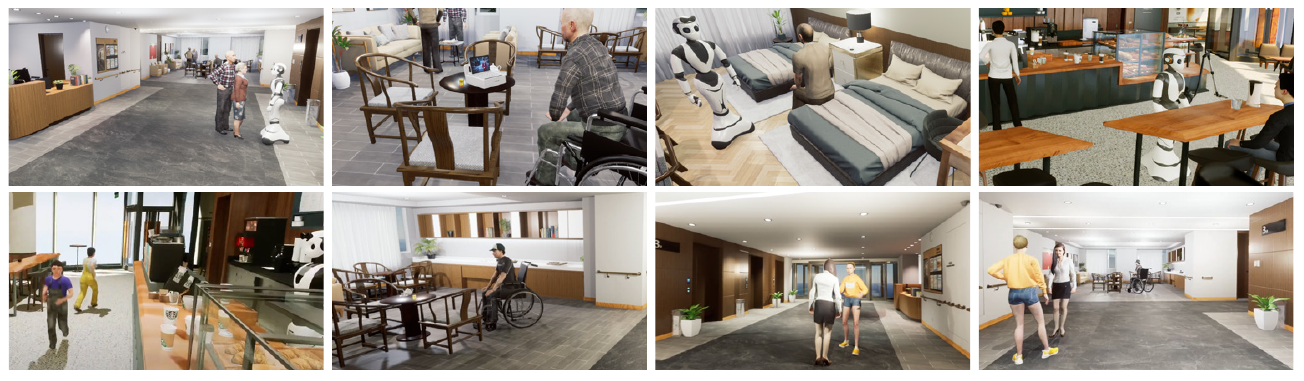}
    \vspace{-2mm}
    \caption{Examples of stationary and walking pedestrians in our simulator.}
    \label{fig:pedestrian}
\end{figure*}

Our simulator supports walking and stationary pedestrians and provides Python API for users to generate and control them, aligning more closely with real-world navigation scenarios. This feature is not included in previous benchmarks and significant for developing navigation methods in complex scenarios. We present some examples in Figure \ref{fig:pedestrian}.

\section{Dataset Details}

\begin{figure}[t]
    \centering
    \includegraphics[width=1.0\linewidth]{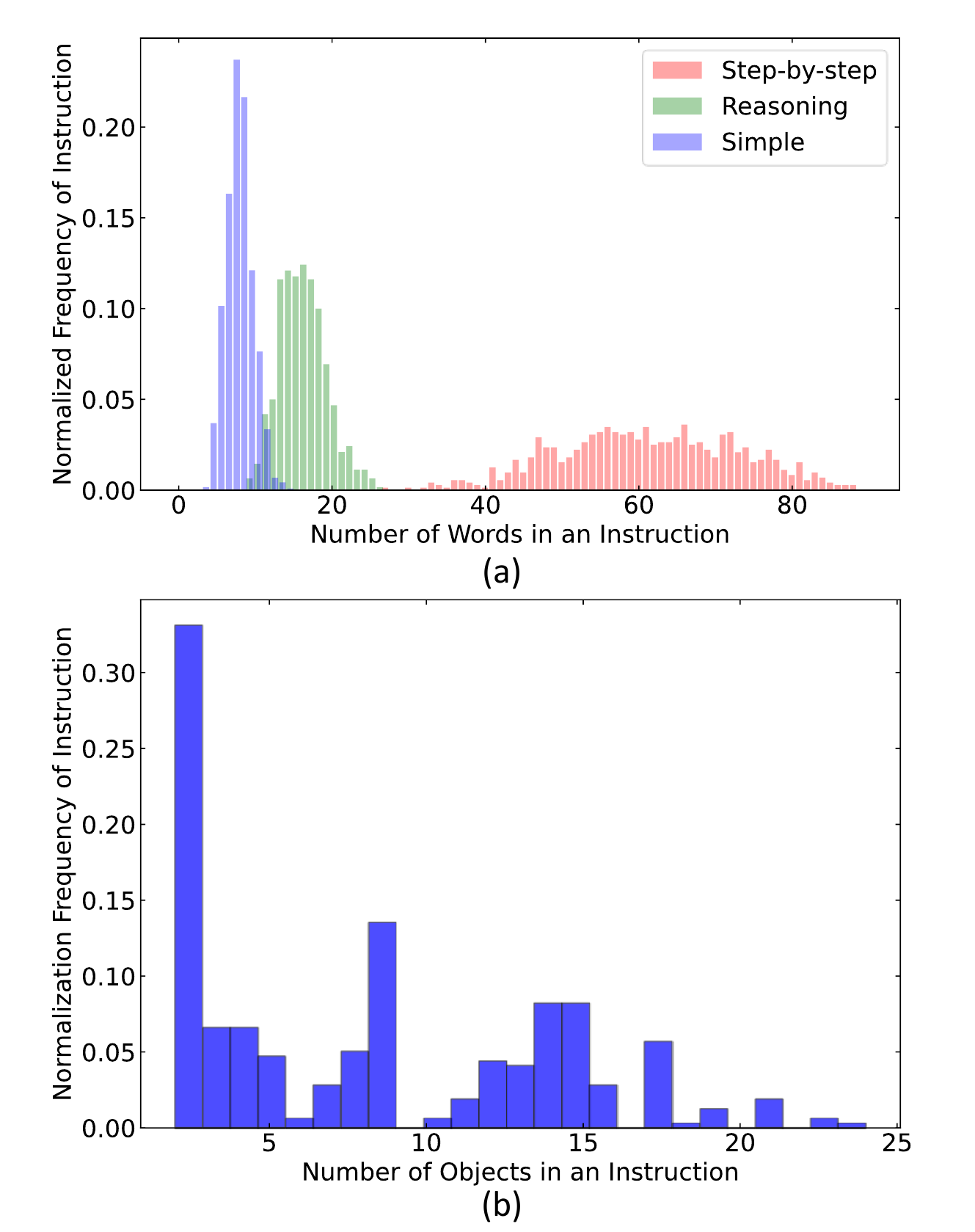}
    \caption{The distribution of the number of words (a) and objects in Step-by-step (b) in each instruction.}
    \label{fig:word_num}
\end{figure}

Figure \ref{fig:word_num} (a) displays the length distribution of the collected instructions for three tasks. It shows that most instructions have 8 $\sim$ 12 words in the simple task, while most instructions have 15 $\sim$ 18 words in the reasoning task. For step-by-step instruction following, most instructions have 50 $\sim$ 70 words.
We also compute the number of mentioned objects in the step-by-step instruction following and its distribution is presented in Figure \ref{fig:word_num} (b). It shows that 32\% instructions mention 2 objects, 14\% instructions mention 9 objects, and around 20\% instructions mention 13 $\sim$ 15 objects.

\begin{figure}[t]
    \centering
    \includegraphics[width=1.0\linewidth]{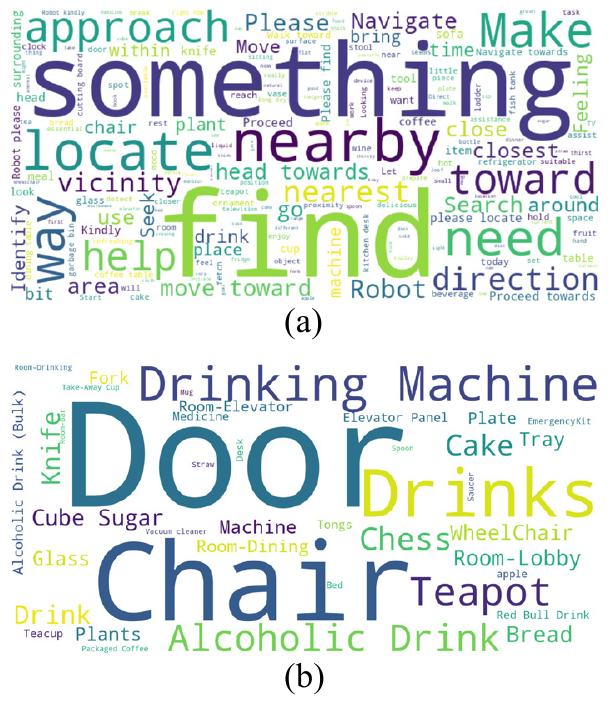}
    \caption{Word cloud of all instructions (a) and target objects in Step-by-step (b) in our dataset. The bigger the font, the more percentage it occupies.}
    \label{fig:word_cloud}
\end{figure}

Figure \ref{fig:word_cloud} (a) presents the relative amount of words used in instructions in the form of the word cloud. It shows that GPT-4 prefers to generate `find', `nearby', and `toward' for navigation. Most instructions involve `something'. For step-by-step instruction following, most instructions involve `door' and `chair'.

\section{Prompt Details}
\subsection{Prompts for Data Collection}
We have to generate instructions for three tasks: goal-conditioned navigation given a simple instruction, completing abstract instruction, and step-by-step instruction following. We detail prompts for different tasks during data collection in the following.

\subsubsection{Goal-conditioned navigation given a simple instruction}

\begin{tcolorbox}[breakable]
As the navigator of a walking robot, your task is to provide clear instructions for it to find specific objects.

When I tell you the name of the item that the robot needs to go to, please output a command to the robot.

For example:

when I say "Bread", you should say "Get close to a loaf of bread", "Find a loaf of bread", "Go to a loaf of bread";

when I say "Teacup", you should say "Please go to find a teacup", "Get to a teacup", "Robot, please find a teacup";

when I say "Drinking Machine", you should say "Seek out a drinking machine in your vicinity", "Get near to a drinking machine  nearby", "Navigate towards a drinking machine”;

when I say "Glass”, you should say "Robot, help me to find a glass", "Make your way to the glass I referred to", "Navigate to a glass nearby”;

Please adhere to the following guidelines:

1. Avoid using phrases such as "for example”, "I can say”, "or”, and "and”;

2. Only provide clear instructions, and additional inquiries are not allowed;

3. Do not refer to any specific location, such as a kitchen or drawing room;

4. Only mention the provided items and avoid mentioning any other objects;

5. Avoid explicitly stating where the robot should retrieve an object from;

6. Use simple language when providing instructions.

Remember, when giving instructions, make sure they are clear enough for the robot to understand which object it needs to go to. The instructions should only require the robot to find and walk toward that specific object. Avoid mentioning any additional requirements beyond walking.

So, give the robot 10 instructions about \{Object\}:
\end{tcolorbox}

\subsubsection{Completing abstract instruction}
\begin{tcolorbox}[breakable]
As the navigator of a walking robot, your task is to provide clear instructions for it to find specific objects.

When I tell you the name of the item that the robot needs to go to, please output a command to the robot.

For example:

when I say "bread", you should say: "I'm hungry and want something to eat.", "I just got off work and haven't eaten yet. Find me something to eat".

Please adhere to the following guidelines:

1. Avoid using phrases such as "for example", "I can say", "or", and "and";

2. Only provide instructions; additional inquiries are not allowed;

3. Do not refer to any specific location, such as a kitchen or drawing room;

4. Only mention the provided items and avoid mentioning any other objects.

5. Avoid explicitly stating where the robot should retrieve an object from.

6. Use less direct expressions

Remember, when giving instructions, make sure they never mention the name of the given object. The instructions should only require the robot to find and walk toward that specific object. Avoid mentioning any additional requirements beyond walking. And NEVER DIRECTLY reference to the name I give.

For example:

I say "Banana", You can say: "Feeling a bit low on energy. Could you fetch something to pick me up?", "Find me something I can eat on the go, without needing a plate or utensils", "Just got back from my run. I want to use a quick natural energy booster", "Find me a snack that's a good source of natural sugars";

I say "Blood pressure monitor", You can say: "I've been feeling a bit light-headed lately. I need to check something important", "I think it's time for a quick health check, but nothing too invasive", "I've been feeling a bit odd lately, and want to use something to check my wellness".

So, give the robot 1 instruction about \{Object\}:
\end{tcolorbox}

\subsubsection{Step-by-step instruction following}

\begin{tcolorbox}[breakable]
Imagine that you are someone who is good at expressing needs. Now, you need to command a robot to follow a certain path and reach a specific object. Remember the following rules:

1. Use everyday expressions as much as possible;

2. Keep your words as simple as possible, don't get too complicated, and focus on the key objects;

3. Do not use expressions like "position 0" to distinguish;

4. Only provide clear instructions; additional inquiries are not allowed;

5. Only mention the items that have been provided and avoid mentioning any other objects;

6. Remember, you can output a command with up to 80 words;

7. Don't concatenate landmarks with verbs that contain location information. Only use verbs like "go to" and "find".

This is the path that you need to command the robot to walk through, recording the name of the nearest object at intervals:

Target name:   \{Target\}

Things around every point of the trajectory: \{Object1\}, \{Object2\}, \{Object3\}.

Now, based on the information above, you need to clearly describe the trajectory given to a navigation robot so that it can follow the path after understanding your description. 

Describe the important information and summarize it into a paragraph of 80 words or less.

Now, give your instructions directly.
\end{tcolorbox}

For this task, we define two types of instructions: instructions providing only landmarks and goals, and instructions providing additional movement sequences. When generating instructions providing additional movement sequences, we provide the movement sequences in the prompt as follows: Actions that the robot should take at each point: \{Action1\}, \{Action2\}, \{Action3\}, $\cdots$.

\section{Self-refinement}

\begin{tcolorbox}[breakable]
\begin{itemize}[topsep=-2pt,leftmargin=8pt]
\item Bring me a little touch of nature that can thrive under indoor care and brighten up the space. (BEFORE: plants, plant, fruit, cloth; AFTER: plants, plant)
\item It's been a tough day, could you find me something to support me while moving around? (BEFORE: chair, wheelchair, armrest, table; AFTER: chair, wheelchair, armrest)
\item Help me find something to neatly pour hot water. (BEFORE: pot, table, teapot; AFTER: pot, teapot)
\item After a long day, it's time to wind down and find a place to rest my eyes. (BEFORE: curtain, chair, sofa, books, bed; AFTER: curtain, chair, sofa, bed)
\item I'm in the mood to choose an outfit; find me an area where I can store clothes neatly. (BEFORE: closet, cloth; AFTER: closet)
\end{itemize}
\end{tcolorbox}

Using LLMs to assess the viability of their own predictions is becoming an increasingly important procedure in problem-solving \cite{shinn2023reflexion,madaan2023self,paul2023refiner,yao2023tree}.
In the task of completing abstract instruction, GPT-4 \cite{OpenAI2023GPT4TR} outputs possible objects for each instruction. Then we use GPT-4 to self-refine its output.
In the above examples, the content inside the brackets represents the desired objects. BEFORE indicates outputs without self-refinement, while AFTER represents outputs with self-refinement. We can conclude that self-refinement improves the initial generation.

\subsection{Prompts for Instruction Parsing}
We design different prompts for decoding instructions as follows.

\subsubsection{Goal-conditioned navigation given a simple instruction}

\begin{tcolorbox}[breakable]
Imagine that you are a very intelligent service robot. 

You receive an instruction from the user: \{Instruction\}

You need to figure out what object the user needs, and then output it. Remember, answer in a short statement, because you can only choose one item.

Your output is:
\end{tcolorbox}

\subsubsection{Completing abstract instruction}

\begin{tcolorbox}[breakable]
Imagine that you are a very intelligent service robot. 

You will receive an instruction from the user, and the only things you can provide to the user are listed as below \{Option1, Option2, $\cdots$\}.

The instruction you have just received is \{Instruction\}.

You need to choose the item you can provide that best fits the user's needs. Remember, answer in a short statement, because you can only choose one item.
\end{tcolorbox}

\subsubsection{Step-by-step instruction following}

\begin{tcolorbox}[breakable]
Given an instruction, you need to extract the landmarks in the instruction and sort them in the order in which they appear in the realistic navigation (not in the order they appear in the instruction).

Requirement 1: Extract all landmarks in the instruction.

Requirement 2: Do not generate landmarks that are not in the instruction.

Requirement 3: Print the landmarks in sequence.

Requirement 4: Don't put anything other than a landmark on the landmark line.

For example: 

Instruction: First, start at the Curtain, then walk along until you see the Plants, and continue heading straight. When you reach the Fridge, take a slight turn towards the right, and just a bit beyond it, you should see the Monitor.

Landmarks:
1. Curtain;
2. Plants;
3. Fridge;
4. Monitor.

Now, you are given an Instruction: \{Instruction\}

Landmarks:
\end{tcolorbox}

\section{Instruction Samples}
We provide some generated instructions for different tasks.

\subsection{Goal-conditioned navigation given a simple instruction}

\begin{tcolorbox}[breakable]
\begin{itemize}[topsep=-2pt,leftmargin=8pt]
\item Locate the nearest monitor.
\item Robot, move towards the knife.
\item Approach the chess set nearby.
\item Walk towards a closet in your vicinity.
\item Seek out any visible book in the area.
\item Please locate and walk to the desired medicine container.
\item Make your way to the nearest table.
\item Seek out an emergency kit in your proximity.
\item Look for a plant and make your way towards it.
\item Proceed to the nearest bed in the area.
\end{itemize}
\end{tcolorbox}

\subsection{Completing abstract instruction}

\begin{tcolorbox}[breakable]
\begin{itemize}[topsep=-2pt,leftmargin=8pt]
\item I'm thinking about doing some screen-based work, find me a device that will help me with that.
\item I'm about to make a sandwich and need a helpful utensial to get the job done efficiently.
\item Feeling inspired to make a cozy, warm meal for tonight. Could you fetch me the key vessel we'll be using to cook it in?
\item In the mood for some strategic fun, can you locate a board game with elegant pieces engraved with  squares on it? 
\item Find me a place where I can enjoy my meal more comfortably.
\item Locate a snack that contains vitamins and is both sweet and refreshing.
\item It's getting a bit chilly, could you find me something cozy to wear?
\item I'm in need of something designed to hold a warm beverage for brewing.
\item I've been on my feet all day, and now I ought to locate a cozy spot to recharge.
\item Feeling a tad warm, could you look for something that keeps food and drinks chilled?
\end{itemize}
\end{tcolorbox}

\subsection{Step-by-step instruction following}

\begin{tcolorbox}[breakable]
\begin{itemize}[topsep=-2pt,leftmargin=8pt]
\item Robot, please start by heading towards the teapot. As you continue, you will pass by a drinking machine multiple times. After that, you'll see an alcoholic drink in the area. Keep going until you find a cake. Make sure you follow the cakes - there will be a few. Your final destination is the tongs. Keep following this path, and you'll reach the tongs. Remember, the keynote of our journey is teapot, drinking machines, alcoholic drink, cakes, and then tongs. Good luck!
\item Robot, I need you to follow this path to reach the Tongs. Begin by moving forward towards the Teapot, continue going straight, passing the Drinking Machine and Alcoholic Drink. Keep moving forward until you reach the Cake and then take a slight left. Proceed forward next to more Cakes, then take a right turn as you keep forward. After that final turn, you will see the Tongs in front of you. Please stop upon reaching the Tongs.
\item Robot, please move forward along this trajectory. As you proceed, you will first notice a monitor nearby. Keep going straight and focus on your path. Then, after approaching the plants, immediately stop. That is your final destination, which we referred to as Plants. Remember to take actions such as Forward and STOP only when necessary. Good luck on your journey!
\item Hey robot, first, I need you to move forward and head towards the door. Make sure they're on your side. Continue along the path, and you'll eventually reach our desired target - the drinks. Keep focused, nice, and simple - locate the door, then finally the drinks.
\end{itemize}
\end{tcolorbox}

\begin{figure*}[t]
    \centering
    \includegraphics[width=1.0\linewidth]{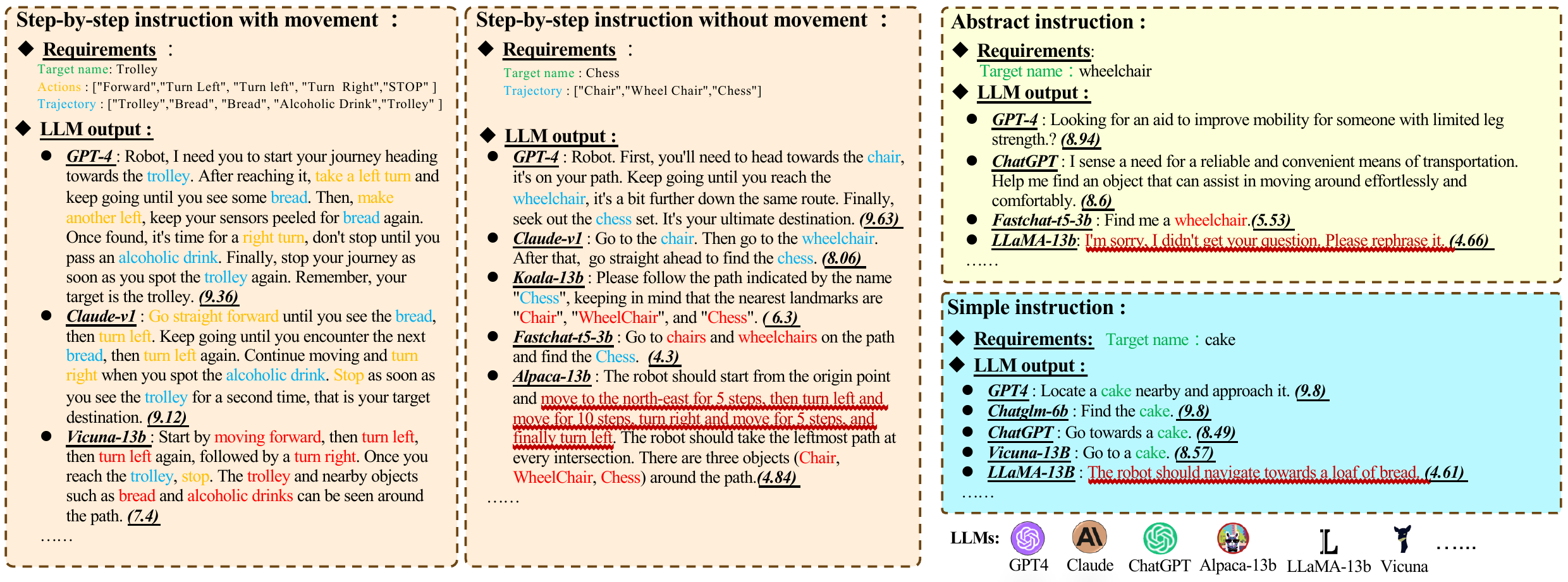}
    \vspace{-2mm}
    \caption{Examples of instruction generated by different LLMs under three different tasks.}
    \vspace{-2mm}
    \label{fig:prompt}
\end{figure*}

\begin{table*}[t]
\centering
\caption{Humans score LLMs on a scale of 0$\sim$10. M represents movement sequences. AVG indicates the average score.}
\resizebox{1.0\linewidth}{!}{
\begin{tabular}{c|cccc|c}
\toprule
 & Simple & Reasoning & Step-by-step w/o M & Step-by-step w/ M & AVG \\
 \midrule
GPT-4 \cite{OpenAI2023GPT4TR}        & \textbf{9.80}   & \textbf{8.94}  & \textbf{9.63}  & \textbf{9.36}  & \textbf{9.43} \\
Claude-v1 \cite{bai2022constitutional}    & 8.26  & 8.39  & 8.06  & 9.12  & 8.46 \\
ChatGPT \cite{ouyang2022training}      & 8.49  & 8.60   & 8.00     & 8.08  & 8.29 \\
Vicuna-13b \cite{vicuna2023}   & 8.57  & 7.42  & 6.80   & 7.40   & 7.55 \\
Vicuna-7b \cite{vicuna2023}    & 8.26  & 7.10   & 7.89  & 6.60   & 7.46 \\
Koala-13b \cite{koala_blogpost_2023}    & 8.57  & 7.30   & 6.30   & 6.44  & 7.15 \\
Chatglm-6b \cite{chatGLM2022}   & \textbf{9.80}   & 6.10   & 5.50   & 5.85  & 6.81 \\
Fschat-t5-3b \cite{vicuna2023} & 9.10   & 5.53  & 4.30   & 5.33  & 6.07  \\
Alpaca-13b \cite{alpaca}   & 8.50   & 4.84  & 4.84  & 5.84  & 6.01  \\
LLaMA-13b \cite{touvron2023llama}    & 4.61  & 4.66  & 2.71  & 5.77  & 4.44 \\
\midrule
AVG          & 8.40 & 6.89 & 6.40 & 6.98 &   -     \\
\bottomrule
\end{tabular}}
\label{tab:score}
\end{table*}

\section{Human Scoring}
We sample 60 instructions generated by different LLMs and ask 100 people to rate their plausibility. As shown in Table \ref{tab:score}, GPT-4 \cite{OpenAI2023GPT4TR} performs best overall. For the simple task, Chatglm-6b \cite{chatGLM2022} performs as well as GPT-4.
The evaluations of the quality of instructions are collected from 100 people, who are undergraduate and graduate students from the university and ages from 18 to 30. They are able to evaluate the dataset accurately.
Examples of some of the higher-scoring and lower-scoring cases are presented in Figure \ref{fig:prompt}.

\section{Qualitative Results}

\subsection{Compare with previous simulators}

\begin{figure*}[t]
    \centering
    \includegraphics[width=\linewidth]{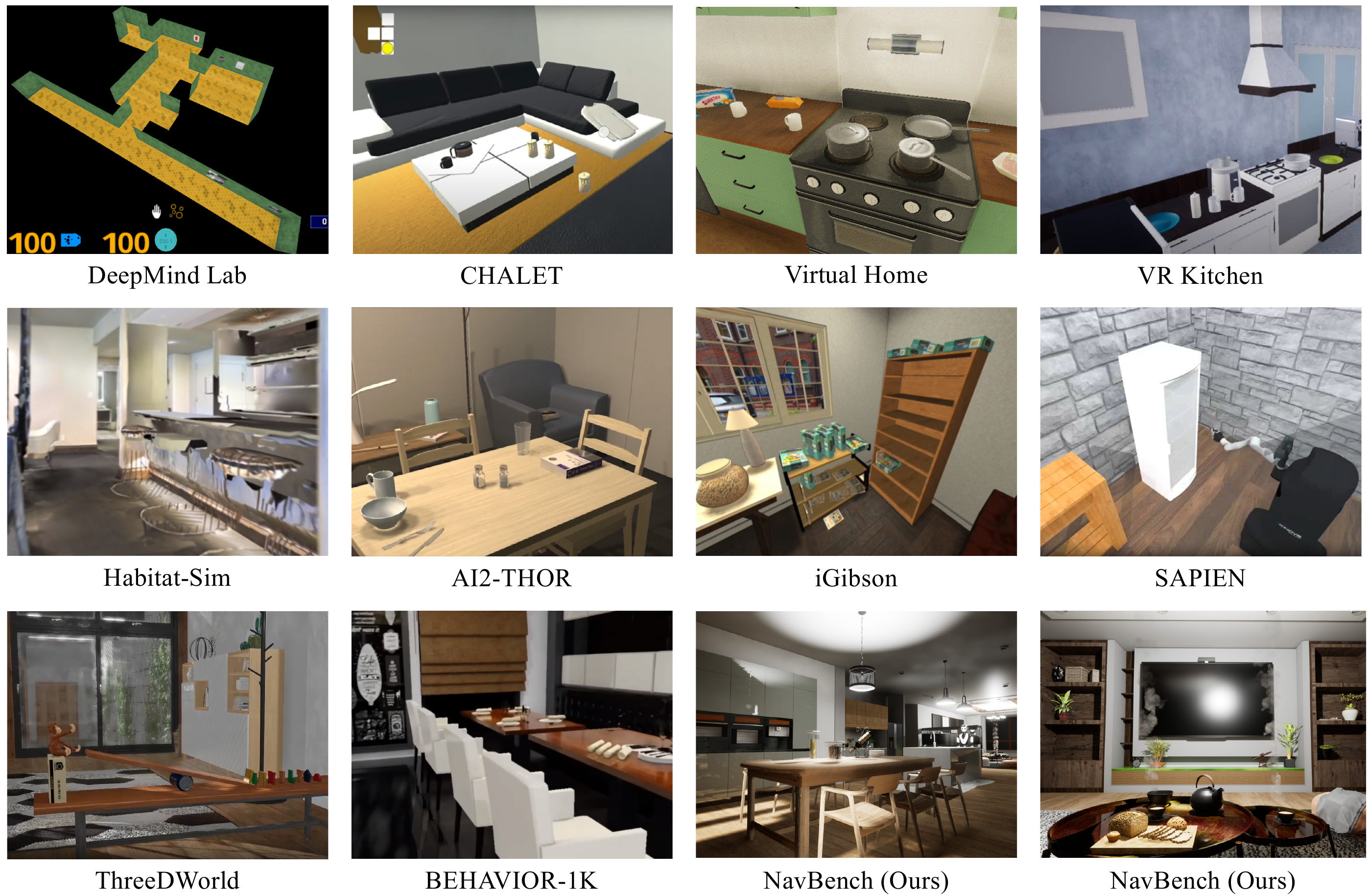}
    \vspace{-2mm}
    \caption{Compare with existing simulators. The last two pictures are from our simulator. Our environment is more elaborate, and the lighting in our simulator is more realistic.}
    \label{fig:simulators}
\end{figure*}

We present qualitative comparisons between different simulators in Figure~\ref{fig:simulators}. These visual comparisons underscore the superior realism and intricacy of our simulator.

\subsection{Environment details}

\begin{figure*}[t]
    \centering
    \includegraphics[width=1.0\linewidth]{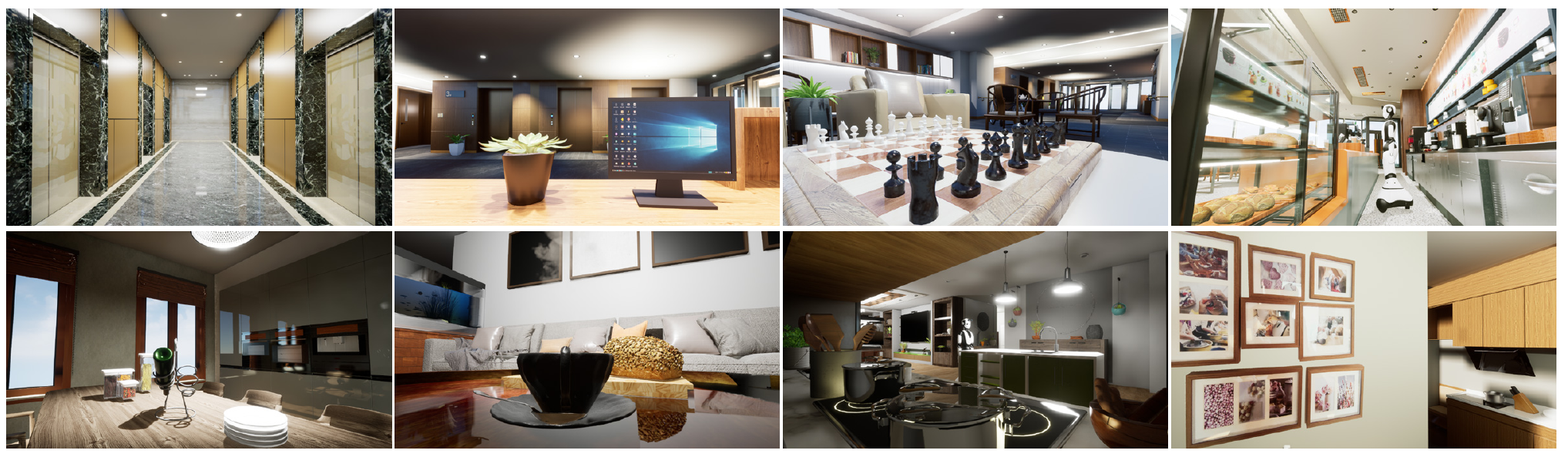}
    \vspace{-2mm}
    \caption{Elaborate environment details in different scenes in our simulator.}
    \label{fig:light}
\end{figure*}

We provide more figures of the environment details in different scenes in our simulator as in Figure \ref{fig:light}. We can see that the illumination, reflections, and shadows are close to the physical world. Besides, our simulator also contains dynamic steam and water, following the principles of the physical world.

\end{document}